\documentclass[12pt]{article}

\usepackage{sbc-template}

\usepackage{graphicx,url,adjustbox}

\usepackage[brazil]{babel}   
\usepackage[utf8]{inputenc}

\title{Human-Level Reasoning? A Comparative Study of Large Language Models on Logical and Abstract Reasoning}

\author{Benjamin Grando Moreira\inst{1}}

\address{Universidade Federal de Santa Catarina (UFSC) - Campus Joinville\\
  Joinville -- SC -- Brasil
  \email{benjamin.grando@ufsc.br}
}

\begin{document} 

\maketitle

\begin{abstract}
  Evaluating reasoning ability in Large Language Models (LLMs) is important for advancing artificial intelligence, as it transcends mere linguistic task performance. It involves understanding whether these models truly understand information, perform inferences, and are able to draw conclusions in a logical and valid way. This study compare logical and abstract reasoning skills of several LLMs - including GPT, Claude, DeepSeek, Gemini, Grok, Llama, Mistral, Perplexity, and Sabiá - using a set of eight custom-designed reasoning questions.
  The LLM results are benchmarked against human performance on the same tasks, revealing significant differences and indicating areas where LLMs struggle with deduction. 
\end{abstract}
     

\section{Introduction}

Large Language Models (LLMs) have recently garnered significant attention as a transformative technology within the field of Artificial Intelligence, particularly due to their demonstrated proficiencies in natural language processing and text generation. Models such as GPT, BERT, and their successors have exhibited considerable efficacy across a spectrum of applications, ranging from automated translation to content synthesis and programming support. The capacity of these models to comprehend and generate text in a coherent and contextually relevant manner signifies a notable advancement, rendering them valuable assets across diverse domains \cite{brown2020language,devlin2019bert}.

However, with the increasing deployment of LLMs, a critical imperative emerges to rigorously evaluate their performance in tasks demanding logical and abstract reasoning acumen. Such tasks are of fundamental importance, as they present challenges that extend beyond mere lexical association, necessitating a deeper understanding and the ability to establish logical interconnections between concepts. A comparative assessment of LLM performance in logical and abstract challenges is, therefore, essential to delineate their inherent limitations and potential capabilities, thereby informing the development of future models capable of surmounting these constraints.

Despite the demonstrated capabilities of LLMs in tasks predicated on natural language processing, such as translation and text generation, their performance in tasks requiring logical and critical reasoning remains suboptimal. Prior research \cite{marcus2019rebooting,bender2021dangers} suggests that while these models adeptly capture statistical patterns within language, they often exhibit inconsistencies in the application of logical principles. This limitation is particularly salient in scenarios where problem resolution hinges on the capacity to comprehend intricate premises and derive valid inferences therefrom.


In light of these considerations, the present study undertakes a comparative analysis of the performance of several LLMs across eight distinct reasoning challenges. The primary objective is to ascertain the extent to which each model can effectively address these challenges. The significance of this analysis lies in its potential to elucidate the relative strengths and weaknesses of each model, thereby providing valuable insights for researchers and developers engaged in efforts to enhance the reasoning capabilities of LLMs.


Existing literature indicates that, notwithstanding recent advancements, numerous models continue to encounter substantial difficulties in tasks necessitating intricate logical reasoning capabilities \cite{marcus2019rebooting,bender2021dangers,Malik2024LostIT}. Addressing these logical challenges constitutes one of the most complex and intellectually stimulating domains within the broader field of artificial intelligence. Such challenges demand that models transcend superficial linguistic comprehension and exhibit the capacity to perform inferences, establish interrelations between concepts, and apply logical rules with precision. This inherent complexity poses a significant impediment to LLMs, which, despite their notable progress, still manifest considerable limitations when confronted with problems requiring sophisticated logical reasoning.

This paper adopts a personalized assessment approach, creating its own challenges and analyzing the responses qualitatively. The paper does not use formal metrics such as Perplexity, BLEU or ROUGE. The assessment is done qualitatively, analyzing whether the LLMs’ answers are correct, whether the judgment is correct (even if the answer is wrong) or whether the answer was not answered. It is an approach that focuses more on analyzing the content of the answer than on statistical statistics. The paper also does not use existing calculation benchmarks such as Winograd Schema Challenge or ARC. Instead, a set of 8 logical reasoning challenges specific to this study is created. This allows for greater control over the type of calculation evaluated, although it is difficult to directly compare with other studies that use standardized benchmarks.

The methodology employed in this study incorporates a customized assessment paradigm, wherein unique challenges were devised, and responses were subjected to qualitative analysis. In contrast to employing standardized quantitative metrics (such as Perplexity, BLEU, or ROUGE), the assessment focused on a qualitative evaluation of the LLMs' responses, determining the correctness of answers, the validity of the reasoning employed (even in instances of incorrect answers), and instances where no response was provided. This approach prioritizes the in-depth analysis of answer content over reliance on statistical metrics. Furthermore, the study diverges from the use of established computational benchmarks like the Winograd Schema Challenge or ARC. Instead, a bespoke set of eight logical reasoning challenges, tailored specifically for this investigation, was created. This affords enhanced control over the nature of the evaluations conducted, albeit at the expense of direct comparability with studies utilizing standardized benchmarks.

The evaluation framework integrates a multifaceted assessment strategy, encompassing: (1) a problem-solving paradigm wherein LLMs are tasked with resolving logical problems; (2) an analysis of response accuracy, involving verification of answer correctness; and (3) a reasoning analysis, focusing on the validity of the judgment employed to arrive at the answer, even in cases where the final response is erroneous. This approach underscores a focus on elucidating the cognitive processes underlying the models' responses.

In total, fifteen distinct LLMs were evaluated in this study, each tasked with addressing eight logical and abstract reasoning challenges. The findings of this investigation underscore the difficulties encountered by several LLMs in responding to the questions posed and reveal notable disparities in performance relative to the application of these challenges to a cohort of human participants.

\section{Logical and abstract reasoning}

Logical and abstract reasoning are central components of intelligence, especially so-called fluid intelligence, which involves an ability to solve novel problems and think flexibly. Logical reasoning refers to the ability to deduce correct conclusions from premises using formal rules of logic. It is a core ability for human cognition and is often used as a measure of intelligence in psychometric tests. Abstract reasoning involves identifying patterns, relationships, and rules in complex or novel information, often without relying on prior knowledge \cite{Yu2021AbstractRV}. 
Logical reasoning and intelligence are closely related but distinct cognitive abilities, with strong evidence of a positive relationship between the two, especially in analytical and mathematical domains. 

Logical reasoning is the process of deriving conclusions from established premises using formal rules and principles to ensure the validity of inferences, and is fundamental for critical analysis, problem-solving and informed decision-making. In the context of AI, the importance of logical reasoning lies in the system's ability to analyze data in a structured manner, formulating informed responses or decisions. 


Abstract reasoning involves understanding and manipulating concepts that go beyond concrete experiences or objects, allowing the individual to identify complex patterns, make generalizations and deal with subtle and symbolic relationships \cite{Mitchell2021}. In AI, this type of reasoning allows recognizing analogies, metaphors and extrapolating information to previously unseen contexts, revealing a more advanced level of cognitive flexibility. Examples include identifying hidden rules in numerical sequences or recognizing analogies between structures in different areas of knowledge.


Despite recent advances, implementing robust logical reasoning in LLMs presents several scientific challenges, including the existence of implicit information, the occurrence of logical errors, as well as difficulty in explaining the response. According to \cite{mirzadeh2024}, LLMs have difficulty in deducing information that is not explicitly present in the training data, compromising their capacity for logical inference in situations that require common sense or tacit knowledge.

\section{Materials and Methods}

In this study, a cohort of fifteen distinct Large Language Models (LLMs) was evaluated. The LLMs included: GPT4.1, GPTo4-mini, GPTo3-mini, Claude 3.7 Sonnet, DeepSeek R1, DeepSeek V3, Gemini 2.0 Flash, Gemini 2.5 Pro, Llama 3.1, Llama 4, Perplexity, Grok 2, Grok 3, Mistral, and Sabiá 3, with the latter representing a Brazilian LLM alternative.

Each LLM was presented with the following initial prompt: “Solve the logical and abstract challenges presented below, and elucidate the reasoning process by which you arrived at your solution.” This prompt was designed not only to assess the accuracy of the responses but also to gain insight into the problem-solving strategies employed by each LLM. It was recognized that instances might occur where the response was incorrect, yet the underlying reasoning was valid. Furthermore, certain problems admitted alternative solution pathways.

To evaluate the LLMs' reasoning capabilities, eight distinct problems pertaining to logical and abstract reasoning were developed. These questions were designed to minimize the reliance on specific factual knowledge, with the exception of the final question, and instead emphasize the ability to engage in logical or abstract thought to identify underlying patterns. The specific intent of each question is detailed in Section \ref{sec:intention}, while the anticipated responses are presented in Section \ref{sec:answers}. The questions administered were as follows:

\begin{enumerate}
    \item Consider the following operations: SUN + 1 = MON; WED + 2 = FRI; MON + 7 = MON. What will be TUE + 2?
    \item One elephant bothers many people, two elephants bother bother much more. Three elephants bother many people, four elephants bother bother bother bother much more. What would be the continuation of the text for five and six elephants?
    \item Imagine a simple encoding, so that the word \textit{Artigo} is encoded as \textit{Bsujhp}. How would the word \textit{Paper} be encoded?
    \item Indicate which alternative represents the solution of the operation 3 + 3 x 5: a) 16; b) 20; c) 30; d) 45.
    \item If January is 17, February is 48, March is 95, then May is? \footnote{The study applied the question in Portuguese and adapted for this paper to the corresponding question in English. The original question is: Se janeiro é 17, fevereiro é 49, março é 95, então maio é?}
    \item Consider the following equivalences: JAN = ENERO; FEV = FEBRERO; JUN = JUNIO. What will SET be equal to?
    \item Consider the following operations: Sunday + 1 = Segunda; Sunday + 2 = Tuesday; Thursday + 6 = Wednesday; Monday + 5 = Sábado; Monday + 4 = Friday; Thursday + 4 = Monday. What is the result for Thursday + 3 and Thursday + 4?
    \item Consider the following operations: 1 + 10 = 3; 10 + 11 = 5; 100 + 111 = 11; 1001 + 11 = 12; 1000 + 1000 = 16. How much will 1000 + 110 be?
\end{enumerate}

The questions were presented sequentially in a chat-based format. To assess the consistency of responses, two instances of each LLM were evaluated. The instance exhibiting the highest number of correct answers was selected for further analysis. While variations in responses were observed, instances of changes in the correctness of the answer were infrequent. Furthermore, the LLMs were not provided with feedback regarding the correctness of their responses, thereby preventing iterative attempts to answer the same question within a given chat session.

For comparative purposes, in addition to the LLM evaluations, the questions were also administered to a cohort of human participants. The objective was to compare the performance of LLMs with that of humans. The participant pool comprised students and professors from higher education programs in quantitative disciplines. A total of 80 human participants (54 students and 26 professors) completed the questionnaire anonymously. Consistent with the LLM evaluations, no supplementary information was provided beyond the questions themselves. All questions were presented in Portuguese for both the human participants and the LLMs.

Finally, for each question, the responses were categorized based on whether the answer was correct, whether the reasoning was correct despite an incorrect answer, or whether no answer was provided. A scoring system was implemented, assigning 10 points for a correct answer, 5 points for correct reasoning, and zero points otherwise.

\subsection{Objectives and Design of Reasoning Challenges} \label{sec:intention}

This section elucidates the rationale behind the formulation of each question and outlines the a priori expectations regarding potential responses. A detailed analysis of observed answer patterns is presented subsequently in Section \ref{sec:answers}.

The first question was designed to assess knowledge of the days of the week. The question deliberately employed abbreviations for the days of the week to avoid explicit prompting. It was hypothesized that both human participants and LLMs would be capable of correctly answering this question.

The second question draws inspiration from a children's song, wherein the sequence of repetitions of the word "bother" is correlated with the number of elephants. Successful resolution of this question necessitates the discernment of a mathematical nuance embedded within the word repetitions. It was anticipated that the alternating pattern of repetitions might pose a challenge for LLMs, although it would likely be readily apparent to human participants.

The third question involves the application of the alphabetical sequence to decode a word using Caesar's Cipher, a well-established and elementary encryption technique. It was expected that both LLMs and human participants would be able to correctly answer this question, given the simplicity of the encoding method.

The fourth question presents a straightforward arithmetic operation; however, the correct answer is deliberately omitted from the provided alternatives. This intentional omission was designed to elicit confusion and serves as the primary objective of the evaluation. Furthermore, an alternative, incorrect result can be obtained if the operation is calculated without adhering to the proper order of operations, potentially inducing an error. It was hypothesized that both human participants (comprising students and professors with expertise in quantitative disciplines) and LLMs would correctly solve the arithmetic operation but might struggle with the absence of a correct answer choice, leading to uncertainty regarding whether to select an incorrect option.

The fifth question is characterized by its complexity, as it does not involve a simple sequence but rather necessitates the separation of digits and the identification of distinct logical relationships associated with each digit. This subtle nuance contributes to the difficulty of the question. Inspired by a social media post, it was anticipated that a subset of human participants would successfully answer the question, whereas LLMs would either fail to do so or attempt to derive an analytical solution.

The sixth question requires the association of abbreviated month names in one language with their corresponding names in another language. It was expected that LLMs would readily identify the linguistic context of the question, whereas human participants might encounter difficulties if they lacked familiarity with the target language (Spanish). It is worth noting that the orthography of month names in Spanish exhibits considerable similarity to that in Portuguese.

The seventh question again involves operations with days of the week but incorporates an alternating pattern of language usage in the results. It was anticipated that human participants would correctly answer the question; however, the language alternation might introduce confusion for LLMs, particularly concerning when to switch between languages.

The eighth question presents an operation involving binary numbers, the result of which must be converted to its decimal equivalent. Successful resolution of this question necessitates the realization that the sums invariably involve values composed exclusively of the digits 1 and 0. It was expected that both human participants and LLMs would find this question challenging, given the limited familiarity with binary numbers and the potential for LLMs to overlook the alternative base due to the decimal representation of the operation's results.


Questions 2 and 5 were designed to assess abstract reasoning capabilities, whereas the remaining questions were structured to evaluate logical reasoning. Among the logical reasoning questions, Question 8 is classified as abductive reasoning, Questions 1, 3, 4, and 7 are classified as deductive reasoning, and Question 6 is classified as inductive reasoning.

\section{Results}

The ensuing discussion presents the observed response patterns, accompanied by an overview of the prevalent errors. Subsequently, quantitative analyses of the responses from both human participants and LLMs are provided, along with a detailed examination of the LLM responses.

\subsection{Expected Responses and Common Errors} \label{sec:answers}

The first question required participants to recognize that SUN is an abbreviation for Sunday, MON represents Monday, WED represents Wednesday, FRI represents Friday, and TUE represents Tuesday. The task involved incrementing the day of the week by a specified number of days. Thus, TUE + 2 should yield THU. A common error was the presentation of the answer as ``Thursday'', which, while reflecting correct reasoning, deviated from the response pattern established in the problem.

In the second challenge, the anticipated response was: ``Five elephants bother many people, six elephants bother bother bother bother bother bother much more''. The underlying pattern dictates that for odd numbers, the word ``bother'' appears once, whereas for even numbers, the word ``bother'' is repeated a number of times equal to the number itself. The challenge lies in discerning this non-trivial correspondence between the number and the number of repetitions, with a frequent error being the repetition of ``bother'' five times for the five elephants.

For the third challenge, the encoding of the word ``Paper'' should result in ``Qbqfs'', with each letter being replaced by its subsequent letter in the alphabet. It is important to note that the differentiation between uppercase and lowercase letters was not considered (humans normally did not differentiate usage, so this criterion was adopted). The most common error, observed in both LLMs and human participants, was the confusion of certain letters within the pattern (e.g., selecting a preceding letter instead of the subsequent one).

In the fourth challenge, the correct result of the calculation is 18, which was deliberately excluded from the provided answer choices. An error related to the order of operations could lead to the answer 30 (option C). Several human respondents correctly performed the calculation but selected one of the available alternatives, a behavior also observed in some LLMs.

The fifth challenge proved to be the most complex. The expected answer for the month of May was 254 (maio, in Portuguese), derived from concatenating the square of the month's position with the number of letters in the month's name. In this case, May is the fifth month ($5^2$ = 25) and has 4 letters (in the Portuguese language), resulting in the concatenation of 25 and 4, yielding 254. The LLMs primarily attempted to identify numerical sequences directly associated with elements of the months, such as the number of letters, the month's position, or the assignment of values to the letters. However, the concatenation of two numerical values proved excessively challenging for the LLMs, despite the simplicity of the individual calculations. The most prevalent solution proposed by the LLMs (47\% of the responses) involved calculating the value from a polynomial function ($7n^2 + 11n $ - $ 1$) or a similar formula, a method employed by 11\% of the professors (the only participant group to utilize this approach). The expected answer was provided by only one LLM (Gemini 2.5 Pro), corresponding to 6.7\% of the LLM responses, and by 31.2\% of the human participants (50\% among professors and 22.2\% among students).

The sixth question proved more challenging for human participants than for LLMs due to the need to recognize the Spanish month names (the antecedent is an acronym in Portuguese, and the consequent is the month in Spanish). The similarity between the orthography of month names in Portuguese and Spanish (e.g., fevereiro and febrero) may have led to the perception of the question as involving letter substitution or deletion.

The seventh question clearly established the relationship with the names of the days of the week in Portuguese and English. However, the underlying logic involved a language switch based on whether the increment was an even number (resulting in English) or an odd number (resulting in Portuguese). Identifying when to alternate the language proved challenging for both human participants and LLMs, although identifying the result of the increment in the days of the week was relatively straightforward.

Finally, the eighth question required the conversion of an operation involving binary numbers into its decimal equivalent. The correct result is 14 ($1000_2$ = $8_{10}$, $110_2$ = $6_{10}$, $8_{10}$ + $6_{10}$ = $14_{10}$; ou $1000_2$ + $110_2$ = $1110_2$ = $14_{10}$).

\subsection{Evaluation of LLM Responses}


A synthesis of the LLM evaluation is presented in Table 1, detailing the results for each of the eight questions ($Q_i$), accompanied by the final score attained by each model. For the purpose of data interpretation, the following conventions were adopted: for Question 2 (Q2), the table indicates whether the response adhered to the prescribed pattern of repetitions; for Question 4 (Q4), the table presents the LLM's response, with the selected alternative indicated parenthetically; and for all other questions, the table displays the LLM's response.


\begin{table}[ht]
\centering
\caption{LLMs Performance Comparison Table}
\resizebox{\textwidth}{!}{
\begin{tabular}{|l|c|c|c|c|c|c|c|c|c|}
\hline
\textbf{LLM} & \textbf{Q1} & \textbf{Q2} & \textbf{Q3} & \textbf{Q4} & \textbf{Q5} & \textbf{Q6} & \textbf{Q7} & \textbf{Q8} & \textbf{Score} \\
\hline
GPT4.1 & QUI & Correct & Qbqfs & 18 & 54 & Septiembre & Sunday Monday & 14 & 60 \\
GPT4o-mini & QUI & Correct & Qbqfs & 18 & 229 & Septiembre & Domingo Monday & 14 & \textbf{80} \\
GPT03-mini & QUI & Correct & Qbqfs & 18 & 229 & Septiembre & Domingo Monday & 14 & \textbf{80} \\
Claude 3.7 Sonnet & QUI & Correct & Qbqfs & 18 (C) & 163 & Septiembre & Domingo Monday & 14 & \textbf{80} \\
DeepSeek R1 & QUI & Correct & Qbqfs & 18 & - & Septiembre & Domingo Domingo & 14 & 60 \\
DeepSeek V3 & QUI & Correct & QBQFS & 18 (C) & 45 & Septiembre & Sunday Monday & 14 & 70 \\
Gemini 2.5 pro & QUI & Correct & Qbqfs & 18 & 254 & Septiembre & Sunday Monday & 14 & 70 \\
Gemini 2.0 Flash & QUI & Incorrect & QBGFS & 18 & 229 & Septiembre & Domingo Monday & 3 & \textbf{45} \\
Llama 3.1 & QUA & Incorrect & Qbqfs & 18 (A) & 229 & Septiembre & Sunday Monday & 4 & \textbf{40} \\
Llama 4 & QUI & Correct & Qbqfs & 18 & 229 & Septiembre & Domingo Monday & - & 60 \\
Perplexity & QUI & Correct & Qbqfs & 18 (A) & 155 & Septiembre & Sunday Monday & 14 & 70 \\
Grok 2 & QUINTA & Incorrect & Qbdfs & 18 & 87 & Septiembre & Sábado Segunda & 14 & \textbf{40} \\
Grok 3 & QUI & Correct & Qbqfs & 18 (C) & 229 & Septiembre & Sunday Monday & 14 & 70 \\
Mistral & QUI & Correct & Qbfsq & 18 & 229 & Septiembre & Domingo Monday & 16 & 45 \\
Sabía 3 & QUA & Incorrect & Qbsfq & 18 & 123 & Septiembre & Domingo Segunda & - & \textbf{25} \\
\hline
\end{tabular}
}
\end{table}

All LLMs correctly answered Question 6, as well as Question 4, with 60\% of the responses to Question 4 indicating that none of the provided options were correct. The majority of the remaining responses to Question 4 selected option C, accompanied by an explanation that the order of operations was disregarded. In Question 3, all LLMs articulated the correct procedure, although some models erred in presenting the final answer. In Question 1, two LLMs provided incorrect answers, with one model (Grok 2) deviating from the expected response format. For Question 8, the LLMs that correctly identified the operation as involving binary numbers provided the correct answer.

The mean score attained by the LLMs was 59.7 points, with three models (GPTo4-mini, GPTo3-mini, and Claude 3.7 Sonnet) achieving the maximum score, representing 20\% of the evaluated LLMs. The poorest performance was observed for Sabiá 3, Llama 3.1, Gemini 2.0 Flash, and Grok 2, despite the availability of more recent versions of the latter three models exhibiting improved performance.

\subsection{Detailed assessment}

With regard to the initial prompt, Mistral, Grok3, Perplexity, Llama 4, DeepSeek V3, Claude 3.7 Sonnet, and Sabiá 3 began by proposing logic puzzles, in contrast to the other models, which awaited explicit questions. DeepSeek V3 and Mistral reiterated the same challenge, pertaining to a scenario involving three switches. Claude 3.7 Sonnet and Llama 4 replicated the three-door challenge, also known as the Monty Hall problem.

In response to the first question, all models correctly identified that the initial operand referred to abbreviations for the days of the week and that the increment corresponded to an increase in the number of days. Diverging from this trend, GPTo4-mini and GPT4.1 assigned numerical values to the days of the week and proposed a computational approach to derive the solution. GPT4.1 suggested a solution based on the summation of these values, whereas GPTo4-mini employed the MOD operator (remainder of the division) in its calculations. Although Gemini 2.0 Flash arrived at the correct answer, it exhibited an error in its reasoning process, subsequently rectifying this in its final response (“TER + 2 = QUA: This means Tuesday + 2 days = Thursday. Answer: TER + 2 = QUI”). A similar scenario transpired with Sabiá 3; however, it did not ultimately achieve a successful resolution (``Adding 2 days to TER (Tuesday), we arrive at QUA (Wednesday). Therefore, the correct answer is: TER + 2 = QUA'').

In the context of Question 2, the models Gemini 2.0 Flash, Sabiá 3, Llama 3.1, and Grok 2 recognized that the word repetition was contingent upon the numerical value; however, they failed to discern the pattern governing the occurrence of repetition in odd values.

In addressing the third question, the LLMs readily mapped the letter increment. GPT4.1 mapped the letters using the ASCII table. DeepSeek R1 assigned numerical values to the letters and initially employed only uppercase letters, subsequently recognizing that only the first letter should be capitalized (a refinement not implemented in DeepSeek V3). In Llama 4, the entire process was elucidated using uppercase letters; however, the final output presented only the first letter capitalized. Perplexity provided a somewhat idiosyncratic, yet ultimately correct, explanation of the process, represented as ``P → Qa → bp → qe → fr → s''. Mistral and Sabiá 3, as previously noted, correctly articulated the necessary substitutions but committed errors in the final presentation of the answer.

All models executed the correct calculation and indicated the potential for an error in the question or answer choices. The following models explicitly stated that there was no valid alternative: GPT4.1, GPTo3-mini, DeepSeek R1, Gemini 2.5 Pro, Gemini 2.0 Flash, Llama 4, Grok 2, Mistral, and Sabiá 3. The models that posited the possibility of an error and provided an answer without considering operator precedence were: GPTo4-mini, Claude 3.7 Sonnet, DeepSeek V3, and Grok 3. Claude 3.7 Sonnet proposed alternative calculations that yielded options B, C, and D, as well as the possibility of option A through approximation of the calculated value. Perplexity and Llama 3.1 also suggested that option A could be correct, contingent upon approximation.

With respect to Question 5, the following elaborations were presented:

\begin{itemize}
    \item GPT4.1 indicated that ``there is no obvious mathematical relationship'' after multiple attempts and endeavored to identify a less apparent relationship, albeit without success. Gemini 2.5 Pro identified the intended pattern, involving the concatenation of the square of the month number and the number of letters in the month’s name, whereas GPT4.1 indicated that the second component of the number was based on the number of letters, and Claude 3.7 Sonnet identified a pattern in the first component of the value, corresponding to the square of the month's position.
    \item DeepSeek was the model that presented the most possibilities, attempting to substitute letters for numbers, match values from the ASCII table, search for a pattern with prime numbers, and relate to Roman numerals; however, it ultimately failed to derive a solution. DeepSeek V3 initially explored several alternatives before formulating an equation that considered the month number multiplied by twice the number of letters plus the month number.
    \item  GPTo4-mini and GPTo3-mini formulated an analytical solution based on a second-degree polynomial function from the outset. Grok 3 attempted to identify arithmetic and geometric progressions before proposing the equation for resolution, mirroring the approach employed by GPTo4-mini and GPTo3-mini.
    \item Gemini 2.0 Flash did not propose an equation but rather posited that the non-linear difference between the differences from the previous month was constant. This approach was also adopted by Mistral, Llama 3.1, and Llama 4. 
    \item Perplexity commenced with the solution alternative proposed by Gemini 2.0 Flash; however, recognizing its non-linearity, it elected to evaluate alternative forms. After attempting to substitute letters for numbers, it explored mirrored value patterns (digit order inversion). Ultimately, the model reverted to the initial alternative but committed an error by skipping the month of April.
    \item Sabiá 3 initially attempted a direct increment, then proposed a solution analogous to that of Gemini 2.0 Flash; however, although it did not omit the month of April as Perplexity did, its solution was not applied to April to derive the value for May, resulting in an incorrect answer.

    \item Grok 2 suggested a solution involving multiplication with prime numbers; however, the explanation lacked coherence and did not, in fact, utilize prime numbers.
\end{itemize}

In addressing Question 6, nearly all models initially identified the language context as Spanish, with the exception of DeepSeek V3, which initially attempted to identify a letter transformation pattern but subsequently recognized the language context.

Regarding the seventh question, the models that answered correctly realized that, when the increment is even, the resulting month should be presented in English; otherwise, in Portuguese (GPTo3-mini, GPTo4-mini, and Claude 3.7 Sonnet). GPT 4.1, Gemini 2.5 Pro, Perplexity, Grok 3, and Sabiá 3 noticed that the language could change, but did not propose an alternative for when this occurs. Llama 3.1 and Grok 2 identified the languages but did not notice the need to alternate their use. DeepSeek presented a rather confusing solution and ultimately got it wrong. DeepSeek V3 assigned numerical values to the days of the week and proposed a calculation using the MOD arithmetic operator (remainder of the division). Gemini 2.0 Flash presented a correct solution, but by coincidence, simply giving one answer in Portuguese and another in English, without correctly identifying when to use one language or the other. Something similar happened with Llama 4 and Mistral, who got it right because they wanted to present the answers in Portuguese, but as the second operation already appeared in the question’s premise and used English, they followed what was established in the premise.

GPTo3-mini, GPTo4-mini, DeepSeek R1, DeepSeek V3, Gemini 2.5 Pro, and Grok 2 directly indicated the application of binary addition and the transformation of the result to its decimal equivalent. GPT4.1, Claude 3.7 Sonnet, and Perplexity initially explored summation options before transitioning to the solution involving binary numbers and decimal conversion, with Claude and Perplexity undertaking multiple distinct attempts initially. Grok 3 attempted to interpret the addition as concatenation, then enumerated the number of 1 digits, before ultimately recognizing the relationship between binary and decimal values. Gemini 2.0 Flash and Llama 3.1 merely attempted to enumerate the number of 1 digits in the operands. Llama 4 initially attempted the solution involving binary numbers and decimal conversion but became confused and explored alternative approaches, such as enumerating the 1 digits of the operands, before abandoning the effort. Mistral commenced by summing the digits, then transitioned to summing the numbers and applying a specific rule; however, this approach was convoluted and did not yield the correct value. Sabiá 3 attempted to sum digits with a constant increment and, upon failing to identify a correct increment value, abandoned the attempt to provide an answer.

In conclusion, it is noteworthy that, even with inputs in Portuguese, DeepSeek R1 reasoned in English in the explanations provided for the questions, presenting only the conclusion of the reasoning and the final answer in Portuguese.

\subsection{Evaluation of Human Responses}

The human response data were stratified into student and professor cohorts due to observed disparities in overall performance between these two groups.

Table \ref{tab:pessoas} presents the percentage of correct responses for each question, incorporating a weighted scoring system that accounted for both the absolute correctness of the response and the validity of the underlying reasoning. As an illustrative example, for the first question administered to professors, 77\% of the responses were definitively correct (i.e., answered WED), while the remaining 23\% demonstrated correct reasoning (i.e., answered Thursday or Thursday). Consequently, the overall success rate was calculated as 88.5\% = (10 * 77\%) + (5 * 23\%).

\begin{table}
    \centering
    \caption{Evaluation of the answers given by humans}
    \resizebox{\textwidth}{!}{
    \begin{tabular}{|c|c|c|c|c|c|c|c|c|c|}
        \hline
        Perfil & Q1 & Q2 & Q3 & Q4 & Q5 & Q6 & Q7 & Q8 & Score \\
        \hline
        Professors & \textbf{88.5\%} & 84.6\% & \textbf{88.1\%} & \textbf{92.3\%} & \textbf{65.4\%} & \textbf{69.2\%} & \textbf{96.1\%} & \textbf{80.7\%} & 83.1 \\
        Students & 80.5\% & \textbf{87.0\%} & 87.2\% & 87.0\% & 22.2\% & 12.9\% & 62.9\% & 64.8\% & 63.1 \\
        \hline
    \end{tabular}
    }
    \label{tab:pessoas}
\end{table}

Regarding Question 4, 46\% of the respondents indicated that they would select alternative C. For Question 5, approximately half of the respondents provided the intended value, while the remaining participants arrived at their answers through analytical methods.

Detailed information regarding the specific problem-solving strategies employed by the human respondents cannot be provided, as this aspect was not evaluated in the present study; only the final responses to each question were collected.

In general, the performance of the professors was markedly superior to that of the students, suggesting that cognitive maturity may be a contributing factor to success in addressing these questions.

As a concluding comparison, Table \ref{tab:pessoasxLLMs} presents a comparative analysis of the accuracy rates for both human participants and LLMs. Although the aggregate performance of the LLMs was generally higher, human participants exhibited superior performance on several individual questions.

\begin{table}[h]
    \centering
    \caption{Evaluation of the answers given by humans and LLMs}
    \resizebox{\textwidth}{!}{
    \begin{tabular}{|c|c|c|c|c|c|c|c|c|c|}
        \hline
        Perfil & Q1 & Q2 & Q3 & Q4 & Q5 & Q6 & Q7 & Q8 & Score \\
        \hline
        LLMs & 80.3\% & 66.6\% & 86.6\% & \textbf{100\%} & \textbf{60.0\%} & \textbf{100\%} & 26.7\% & 66.6\% & 73.4 \\
        Humans & \textbf{83.1\%} & \textbf{86.2\%} & \textbf{87.5\%} & 88.7\% & 36.2\% & 31.2\% & \textbf{73.7\%} & \textbf{70.0\%} & 69.6 \\
        \hline
    \end{tabular}
    }
    \label{tab:pessoasxLLMs}
\end{table}

\subsection{Conclusions and future works}


This study has presented a comparative analysis of the logical and abstract reasoning capabilities of fifteen distinct Large Language Models (LLMs) alongside a cohort of human participants. The results reveal a nuanced landscape of LLM performance, highlighting both notable strengths and persistent limitations.

While LLMs demonstrated proficiency in certain tasks, such as identifying patterns in well-defined problems and executing straightforward calculations, they frequently struggled with questions requiring nuanced understanding, abstract thought, and the integration of disparate information. Notably, LLMs often exhibited difficulty in adapting to unconventional problem formats, such as the deliberate omission of the correct answer choice or the need to switch between languages based on a hidden rule. Human participants, particularly those with expertise in quantitative disciplines, generally outperformed LLMs on tasks requiring abstract reasoning and the application of common-sense knowledge.

These findings underscore the continued gap between artificial and human intelligence, particularly in the realm of higher-order cognitive functions. While LLMs have made significant strides in natural language processing and pattern recognition, their ability to truly ``reason'' remains limited by their reliance on statistical correlations and their lack of genuine understanding.

Furthermore, future studies should explore the impact of different prompting strategies. Some models possess Reasoning capabilities (also known as Chain-of-Thought), but this was not utilized because not all models have this feature. However, it represents an important evaluation for future work. Also, for the fifth question, in the models that proposed an analytical solution (resulting in 229 or 45), it would be interesting to inform them of the value for the month of April and verify the change in the solution approach, given that the solution indicated by the models does not align with the value of the informed month.




\bibliographystyle{sbc}
\bibliography{sbc-template}

\end{document}